\documentclass{article}    
\usepackage{hyperref}

\title{Pragmatic constraints and pronoun reference disambiguation: the possible
and the impossible}

\author{Ernest Davis  \\
Dept. of Computer Science \\
New York University}

\begin{document}
\maketitle

\begin{abstract}
Pronoun disambiguation in understanding text and discourse often requires
the application of both general pragmatic knowledge and context-specific
information.  In AI and linguistics research, this has mostly been studied
in cases where the referent is explicitly stated in the preceding text nearby.
However, pronouns in natural text often refer to entities, collections,
or events that are only implicitly mentioned previously; in those cases the
need to use pragmatic knowledge to disambiguate becomes much more acute and
the characterization of the knowledge becomes much more difficult. 
Extended literary texts at times employ both extremely
complex patterns of reference and extremely rich and subtle forms of
knowledge. Indeed, it is occasionally possible to have a pronoun that is 
far separated from its referent in a text. In the opposite direction, 
pronoun use 
is affected by considerations of focus of attention and by formal constraints 
such as a preference for parallel
syntactic structures; these can be so strong 
that no pragmatic knowledge suffices to overrule them. 
\end{abstract}

\section{Semantics, context, and pronoun disambiguation}
\label{secSemantics}
It is well known that pronoun disambiguation in understanding text and 
discourse often requires the application of both general pragmatic knowledge
and context-specific information (Rohde, 2018).

The point is vividly illustrated by Winograd schemas (Levesque, Davis, and 
Morgenstern 2012) such as the well-known example,
\begin{quote}
The trophy doesn't fit in the brown suitcase because it is too big. \\
The trophy doesn't fit in the brown suitcase because it is too small.
\end{quote}
In the first sentence the pronoun {\it it} must refer to the trophy; in the
second, {\it it} must refer to the suitcase. The sentences differ only
in the final words {\it big} and {\it small}, so that distinction, and 
the relation
of the size of two objects to the possibility of fitting one inside the other,
must be determining the referent of the pronoun. 

Similarly, consider the pair of sentences
\begin{quote}
Joan made sure to thank Susan for all the help she had given.    \\
Joan made sure to thank Susan for all the help she had received.   
\end{quote}
In the first sentence the referent of {\it she} is Susan, whereas in the second
the referent is Joan, reflecting the fact that someone who receives
help thanks someone who gives it.

Current natural language processing 
technology is able to achieve high levels of success on collections
of these kinds of examples (Kocijan et al. 2022) by using the patterns of
word associations extracted by deep learning networks as proxies for this
kind of underlying knowledge. But there is no question that people, in 
understanding these texts, are drawing on a deep knowledge of the world
(Lake and Murphy, 2020) and, I would argue, little doubt that, in the long
run, truly intelligent AI systems will need to do likewise.

The above sentences can be expanded to make the issues of pronoun disambiguation
more complex while keeping them well-formed and intelligible.

\begin{quote}
I tried packing the trophy in the brown suitcase, but it was too large,
and I couldn't close it. So instead I'll get the red suitcase, because it
should be large enough that it will fit in it.

Joan made sure to thank Susan for all the help she had given her when she
had been in trouble. Her generosity and sympathy had made an enormous
difference to her.
\end{quote}

More complex examples are possible:
\begin{quote}
Mary White called Susan Brown to let her know that her daughter 
Jennifer had come over after school to the Whites' house to play with their 
own daughter Edith. She thanked her for calling, because she had really 
started to wonder why she hadn't come home.
\end{quote}

Both AI and linguistic analysis have focused on short texts like the above,
for ease of analysis; but in extended texts, the complexities of reference and
the subtlety of the knowledge needed for disambiguation can be much greater.
For example,
the following passage occurs in part 2, chapter 2 of Jane Austen's {\em Emma}.
Four female characters are involved: Emma Woodhouse, Jane Fairfax,
Jane's aunt Miss Bates, and Miss Bates' mother. The dinner party also includes
at least Emma's father and Mr. Knightley. I have put personal pronouns 
in bold face, for emphasis.\footnote{I am told that readers who encounter this
out of the context of the novel find some of these pronouns hard to resolve.} 

\begin{quote}

Upon the whole, Emma left {\bf her} [Jane Fairfax]  
with such softened, charitable feelings, 
as made {\bf her} look around in walking home, and lament that Highbury 
afforded no young man worthy of giving {\bf her} independence; nobody that 
{\bf she} could wish to scheme about for {\bf her}. 

These were charming feelings  --- but not lasting. Before {\bf she} had committed 
herself by any public profession of eternal friendship for Jane Fairfax, 
or done more towards a recantation of past prejudices and errors, than 
saying to Mr. Knightley, ``{\bf She} certainly is handsome; {\bf she} 
is better than 
handsome!'' Jane had spent an evening at Hartfield with {\bf her} grandmother 
and aunt, and every thing was relapsing much into its usual state. Former 
provocations reappeared. The aunt was as tiresome as ever; more tiresome, 
because anxiety for {\bf her} health was now added to admiration of {\bf her} 
powers; 
and {\bf they} had to listen to the description of 
exactly how little bread 
and butter {\bf she} ate for breakfast, and how small a slice of mutton for 
dinner, as well as to see exhibitions of new caps and new workbags for 
{\bf her} mother and {\bf herself}  and Jane's offences rose again. {\bf They} 
had music; Emma was obliged to play; and the thanks and praise which necessarily 
followed appeared to {\bf her} an affectation of candour, an air of greatness, 
meaning only to shew off in higher style {\bf her} own very superior performance. 
{\bf She} was, besides, which was the worst of all, so cold, so cautious!
\end{quote}

Some of these disambiguations are straightforward, some quite 
subtle.\footnote{Also subtle is the
identification of Jane as the speaker of those thanks and praise that seemed
to Emma an affectation of candor. Presumably everyone at the dinner party
gave Emma thanks and praise for her playing, but it was only Jane's that she
found affected.} Teasing out all the forms of knowledge that are used in all
these disambiguations would require a lengthy analysis. However, I note
that the fact that the narrative is given from Emma's point of view ---
a fact about narrative structure --- is important in disambiguating ``appeared
to her'' in the second to last sentence.

\section{Antecedents that are entirely missing}
\label{secMissing}
There are also a number of cases where a third-person
pronoun can be used with no antecedent
noun phrase.  Some of these are natural, some are literary. Cases II and III are 
common and established enough that they are listed in the Oxford English
Dictionary as distinct meanings of {\it they}.

I. A pronoun may refer to an action or event that appears as a verb phrase.
E.g. in ``John hiked for three hours but it tired him out'' the word {\it
it} refers to the hike.

II. {\it They} can mean people in general or some implicit collection of people.
Sometimes {\it they} is just equivalent to ``Some people'' as in ``They 
say that a shot of whiskey is good for a cold.'' But often it refers to
a specific collection of people which much be inferred from context. For
instance, in ``I went to the hospital but they told me to go home'', 
{\em they} refers to the hospital staff. 
In the earlier quotation from {\em Emma\/}, in the phrase ``they had to listen''
the word {\it they} refers to everyone at the dinner other than Miss Bates 
herself;
in the phrase ``They had music'', it refers to everyone at the dinner. Neither
group is otherwise named here by a noun phrase.

This is possible even when there are other possible antecedents and other
occurrences of {\it they}: ``Jack and Otto went to the hospital, but they told
them to go home.''

However, there are limits to this; it would be a stretch to say, 
with no antecedent  \\
\hspace*{2em} (*) I went to the park and 
they were all enjoying the pleasant spring weather. \\
It would be impossible to say \\
\hspace*{2em} (*) Between January 2020 and March 2021, 
more than 2,500,000 of them 
died of Covid-19. \\
(I use the linguists' convention of marking anomalous sentences with
an asterisk.) 

In the Ira Gershwin lyric, 
``The way you wear your hat / The way you sip your tea / The memory of all
that / No, no, they can't take that away from me,'' the phrase {\it they can't}
means ``No one can''. 

The first-person plural pronoun {\it we\/} and the second-person
pronoun {\it you\/}, which generally have no antecedent, 
raise similar issues. {\it We} refers to
some collection of people that includes the speaker; {\it you\/} refers to
some collection of people being addressed; but which particular collection has
to be determined from context, and indeed is often deliberately vague.

III. The Oxford English Dictionary gives as one definition of the word
{\it they}: 
``People in authority collectively, regarded as impersonal and oppressive''
(OED). A couple of illustrative quotes: \\ 
``English political thinking is much governed by `they'. `They' are the
higher-ups, the mysterious powers.'' --- George Orwell. \\
``They really have to do something about all the potholes in this
city.''

IV. The protagonists of a poem or short story (e.g. ``Stories'' by 
I.L.~Peretz) may be denoted throughout by {\it he} and {\it she} for literary
effect. An extreme example is the novel, {\em Wolf Hall\/}, by Hilary Mantel,
in which the protagonist, Thomas Cromwell, is always denoted as {\it he} 
and never
named; but this at times becomes hard for the reader to follow.

V. There is a literary trope in which characters or the narrator
use pronouns without need for antecedent 
to refer to powerful, malevolent figures whose name is
either unknown or dangerous to speak: for instance, human beings in {\em Bambi},
or Queen Ayesha in Rider Haggard's novel {\em She}. In J.R.R. Tolkien's 
{\em The Two Towers}, Pippin says ``Then {\it he\/} came'' (italics in
the original) without antecedent
to mean Sauron in describing their encounter through the Palantir.
In Conan Doyle's {\em The Valley of Fear,} Porlock writes, ``Dear Mr. Holmes,
I will go no further in this matter. It is too dangerous --- he suspects me,''
{\it he\/} being Professor Moriarty.
In the case of {\it they} this often combines with (III):
``I suppose you mean we ought to spend all our time sucking up to Them, 
and currying favour, and dancing attendance on Them like you do.'' (C.S. Lewis,
{\em The Silver Chair}). (Here {\em Them} is a clique of bullies at a school.) 
In this usage, the pronoun tends to be typographically
marked: put in quotation marks, or italicized, or capitalized, presumably
to be read aloud to a particular emphasis --- the quote from
{\em The Valley of Fear\/} is an exception. It is more common in direct 
quotation than in narration --- Haggard's {\em She\/} is an exception.

\section{Demonstrative pronouns referring to a category}

Somewhat similar to the ambiguity of ``we'' and ``they'' discussed above,
it is possible to use the pronouns ``this'' and ``that'' in such a way that,
though the grammatical antecedent is a specific individual or event, the 
actual semantics is some category that includes the individual.\footnote{Thanks
to Doug Hofstadter for bringing this to my attention.} Such cases are often 
ambiguous, and if they can be disambiguated, they require pragmatic constraints.

I have not found any examples in literature, but here is an artificial example.

\begin{quote}
Joe was way behind on the programming project for his operating systems course,
and it was due the next day, so he turned off his phone and stayed up all night
finishing it. He had never done that before.
\end{quote}

What exactly had Joe never done before? 
\begin{itemize}
\item[1.] Stayed up all night? 
\item[2.] Stayed up all night doing school work?
\item[3.] Stayed up all night doing a programming project (either for a course
or for fun)?
\item[4.] Stayed up all night doing a course for this particular class? or
doing this specific project?
\item[$1'$--$4'$.] Turned off his phone and [1--4]?
\end{itemize}

\section{Excluded forms}
\label{secExcluded}
There are limits on the ways in which a pronoun is allowed to connect
to a referent, even when pragmatic constraints decisively rule out any other
possible referent, as the examples in table~\ref{tabImpossible} demonstrate:

\begin{table}[h]
\begin{quote}
1. ? The trophy doesn't fit in the suitcase, because the trophy is an awkward
shape and it is too small. \\
2. * When Joan broke her leg, Susan had given her a lot of help, but Mary had 
not bothered to ask after her. She had been very generous. \\
3. * Ann has two sons. Barbara has no children. 
Her children are in elementary school.  \\
4. * Robert woke up at 9:00 AM, while Samuel woke up at 6:00 AM, so he had
more time to get ready for school. \\
5. * Robert woke up at 9:00 AM, but Samuel woke up at 6:00 AM, so he had
less time to get ready for school. \\
6. * The child was screaming after the baby bottle and toy fell. Since he
was full, it stopped his crying.
\end{quote}
\caption{Pronoun references in which pragmatic constraints do not disambiguate}
\label{tabImpossible}
\end{table}

(Examples 4 and 6 are from Sakaguchi et al., 2020.) Semantically, {\it it}
in (1) must refer to the suitcase; {\it she} in (2) must refer to Susan;
{\it her} in (3) must refer to Ann; 
{\it he} in (4) must refer to Samuel; 
{\it he} in (5) must refer to Robert; 
and {\it it} in (6) must refer to bottle ---
at least, that is the consensus of the crowd workers who created (6) and 
the other crowd workers who endorsed it. To my ear, 2, 3, 4, and 6 
are completely 
impossible; and 1 and 5 is at best extremely unnatural.\footnote{I am just using my
own ``linguistic intuition''; I have not checked this with empirical studies.}

Note that, (4) and (5) are fine with the conjunctions reversed:
\begin{quote}
$4^{\dagger}$ 
Robert woke up at 9:00 AM, but Samuel woke up at 6:00 AM, so he had
more time to get ready for school. \\
$5^{\dagger}$ 
Robert woke up at 9:00 AM, while Samuel woke up at 6:00 AM, so he had
less time to get ready for school. \\
\end{quote}

Adding a number-agreement constraint fixes (1), and adding 
gender-agreement constraints fixes (4)
and at least helps in (5), but 
neither gender nor number constraints help in (2) or (3) (table~\ref{tabGender}).

\begin{table}[h]
$1'$.  The trophies don't fit in the suitcase, because they are an awkward
shape and it is too small. \\
$2'$. * When Joan broke her leg, Susan had given her a lot of help, but Bob had 
not bothered to ask after her. She had been very generous. \\
$2''$. * When Joan broke her leg, the Millers had given her a lot of help, 
but Bob had not bothered to ask after her. They had been very generous. \\
$3'$. * Ann has two sons. Fred has no children. 
Her children are in elementary school.  \\
$3''$. * Ann and David have two sons. Barbara has no children. Their children
are in elementary school. \\
$4'$. Robert woke up at 9:00 AM, while Sarah woke up at 6:00 AM, so she had
more time to get ready for school. \\
$5'$. ? Robert woke up at 9:00 AM, but Sarah woke up at 6:00 AM, so he had
less time to get ready for school.
\caption{Pronoun references with gender and number constraints. These do not
always disambiguate.}
\label{tabGender}
\end{table}

The question is, why are sentences (1-6) excluded whereas the examples discussed
earlier --- much more complex in structure, much less decisive in their
pragmatic constraints --- acceptable?

It seems plausible to conjecture that to some extent the issue here is one of
narrative focus.  The conjecture would be that there a rule that the referent of a 
pronoun must be sufficiently in focus in the narrative. Because of the syntactic
form of the texts, in sentences 1-6 the referents that are pragmatically possible
(the suitcase, Susan, Ann, Samuel, Robert, and the bottle, respectively) are
out of focus, and therefore the reference is impossible.

However, this is obviously largely explanation of {\em ignotum per ignotis};
without a mechanism for determining what the objects in focus at any point 
in a text,
this raises more questions than it answers. Why does the semantically vacuous 
difference between (4) and (5) in table~\ref{tabImpossible}
so entirely change the focus? How is it that, in the passage from {\em Emma\/},
Emma, Jane, and Miss Bates all remain sufficiently in focus that a pronoun can
refer to any of them? If there is such a rule of focus involved in the variants 
of sentence (1), why do the constraints of number agreement easily suffice to 
overrule it, while the pragmatic constraints do not?

Kehler (2015)
similarly gives a number of examples where the form of the 
sentence either forces a pragmatically implausible reading or 
conflicts with pragmatics in a way that makes the sentence unintelligible.
(Kehler carried out tests with human subjects to confirm his intuitions.)

\begin{quote}
 (a) The demonstrators were denied a permit by the city council because they
feared violence. \\
(b) The city council denied the demonstrators a permit because they felt
strongly that the best way to draw  attention to current political issues is to
advocate violence.\\
(c) Norm lent his car to his brother's girlfriend. He doesn't own one. \\
(d) Margaret Thatcher admires Hillary Clinton, and George W. Bush absolutely
worships her.
\end{quote}

In (a) readers generally interpret {\it they} as referring to the demonstrators.
In (b) readers ``garden path": they initially interpret {\it they} 
as meaning ``the city council" and then consciously backtrack and 
correct when they reach the end of the sentence. In (c) readers end up 
confused rather than deciding that {\it he} refers to the brother. In (d) 
readers interpret {\it her} as meaning Clinton rather than Thatcher, despite 
the fact that, when the example was devised,
Thatcher was much more plausible. In cases (a), (b), and (d), readers 
are letting
syntactic parallelism trump pragmatic plausibility. In (b), in contrast
to Winograd's original example, presumably the distance between 
{\it they} and the
disambiguating ``advocate violence'' is so large that readers cannot suspend
judgement, so they go with the syntactically parallel reading and then withdraw
it. (Why (b) leads to garden-pathing whereas (c) and (d) just lead to puzzlement
is not at all clear.)

Rohde (2018) proposes a Bayesian framework for pronoun reference disambiguation:
probabilities based on the form of the sentence are combined with probabilities
derived from pragmatic considerations as dictated by Bayes' law. The 
cases that we have considered are not {\em inconsistent\/} with a Bayesian 
framework --- very little is, if you adopt the right framework and set the
priors properly (Davis, 2013) --- but they are certainly an awkward fit. One
would have supposed that the assumption that a sentence is meaningful has a
strong prior; that the pragmatic constraints in these sentence have very 
strong priors (e.g. the probability that Barbara both has no children and that her
children are in elementary school is zero); and that the probabilities associated
with these rather 
nebulous considerations of focus would be comparatively indeterminate.
Apparently, that is not the case.

\section{Remote pronoun antecedents in literature}
\label{secRemote}
In literature\footnote{It would probably 
be frowned on in other forms of writing.} it is occasionally possible
to use pronouns whose nearest antecedent occurs much earlier, outside the
categories of section~\ref{secMissing}.

There is a remarkable instance in Dickens' 
{\em David Copperfield}. In chapter 31,
Steerforth seduces Emily and flees with her. Steerforth is mentioned again,
in passing at the beginning of chapter 34 (p.~474 in my copy). Steerforth
is unmentioned in chapters 35-39. In chapter 
40, Mr. Peggoty meets with David Copperfield and describes his search in 
Europe for Emily; in this entire chapter there is only one mention of 
Steerforth, by a pronoun: ``{\em He} was nowt to me now.'' (p.~566; italics
in the original).
The next mention after that is in chapter 46 (p.~645):
\begin{quote}
``Pray, has this girl been found?

``No.''

``And yet she has run away!'' \ldots

``Run away?'' I repeated

``Yes! From him,'' she said with a laugh.
\end{quote}

Another example is from Dickens' {\em Bleak House:}
\begin{quote}
``Jo," repeats Allan, looking at him with attention, ``Jo! Stay. To 
be sure! I recollect this lad some time ago being brought before the coroner."

``Yes, I see you once afore at the inkwhich," whimpers Jo. ``What of that? 
Can't you never let such an unfortnet as me alone? An't I unfortnet 
enough for you yet? How unfortnet do you want me fur to be? I've been 
a-chivied and a-chivied, fust by one on you and nixt by another on you, 
till I'm worritted to skins and bones. The inkwhich warn't MY fault. 
I done nothink. He wos wery good to me, he wos; he wos the only one I 
knowed to speak to, as ever come across my crossing.''
\end{quote}

{\it He} in the last sentence refers to the subject of the coroner's inquest, 
Captain Hawdon, who has not
been mentioned since chapter 40.  Hawdon is an important figure in the back-story,
being the unknown
father of the first-person protagonist Esther Summerson, but a rather
minor figure in the novel as a whole; and many male characters are mentioned
between chapters 40 and 46. 

However, the identification presents no difficulty, either for Allan Woodcourt
or for the reader. The context of the inquest would probably suffice, but
additionally, in chapter 11, which describes the inquest, Jo uses the identical
phrase ``He wos wery good to me, he wos,'' there unambiguously of Hawdon, and 
Dickens marks it by repeating it at the end of the chapter.

A third example is from {\em At Swim-Two-Birds} by Flann O'Brien:
\begin{quote}
On the morning of the third day thereafter, said Finn, he was flogged until
he bled water.
\end{quote}

The referent of {\it he} is a hypothetical ``man who mixed his utterance 
with the
honeywords of Finn''. Finn is continuing his own train of thought,
which in the
novel has been interrupted for two pages by an unrelated conversation 
between two other people. The distance is much shorter, but the
referent is even more minor, and, unlike the quotes from Dickens, is 
completely unrelated to the immediate context.

It can be objected that all three 
is in direct quotation, and anything is 
permitted in direct quotation. However, at least with the quotation
from {\em Bleak House,} if we modify the passage to take it
out of direct quotation, the reference remains clear, though certainly
the literary quality suffers:

\begin{quote}
``Yes, I see you once afore at the inkwhich," whimpers Jo. ``What of that? 
Can't you never let such an unfortnet as me alone? An't I unfortnet 
enough for you yet? How unfortnet do you want me fur to be? I've been 
a-chivied and a-chivied, fust by one on you and nixt by another on you, 
till I'm worritted to skins and bones. The inkwhich warn't MY fault. 
I done nothink.'' Jo continues to whimper that he had been very good to Jo;
he was the only person who came across Jo's crossing that Jo had known to 
speak to.
\end{quote}

(It would be hard to recast the quotes from {\em David Copperfield\/} 
or {\em At Swim-Two-Birds} in this way.)

More importantly for our purposes, in all three cases, the reader identifies
the reference, and the expression does not seem unnatural. They
are not problematic in the manner of the starred sentences in 
section~\ref{secExcluded}.

Narrative focus is clearly relevant in the quotes from {\em David Copperfield},
where Emily is the subject, and in {\em Bleak House}, where the inquest is
the subject. The immediate context is entirely irrelevant in {\em At 
Swim-Two-Birds}; the effect is startling and humorous. That is a much smaller
gap, so the reader still has in mind Finn's previous conversation.

The most striking kind of example of this kind would be the use of {\it he\/} or
{\it she\/} by  the narrator
a significant distance from the last mention of the person, in some 
fairly straightforward\footnote{{\em At Swim-Two-Birds} is anything but
straightforward narrative, and {\em Bleak House} has a somewhat unusual
narrative form,
but their complexities are not relevant to this issue.}
narrative text.  
I expect that such examples exist, but there is
no easy way to search for them. 

\section{Conclusion}
These kinds of phenomena have been studied in linguistics, but, as far as 
I have found, only in disambiguating third person pronouns, only with short
texts, and only when the possible antecedents are given explicitly.
It has
been unstudied in AI studies of text interpretation, because those are 
concerned  with 
texts that do arise, not with explaining why impossible texts cannot arise;
and in AI studies of text generation, because a generator that is conservative
in its use of pronouns can easily avoid the issue. It does become significant
when one is involved in creating a data set of hard pronoun-resolution problems,
because then one has to deliberately skate close to the rather fine
line that separates pronoun usages that are difficult for an AI to resolve 
from pronoun usages in these ways; and it is important that all the texts in
a dataset are clearly proper usage (Kocijan et al. 2022).
The phenomenon is in any case a curious one,
and an adequate explanation might well shed some light on larger questions
of how pronouns are used and understood.

\subsection*{Acknowledgements}
Thanks to Yuling Gu, Doug Hofstadter, 
Andrew Kehler, and Leora Morgenstern for helpful 
discussions.

\subsection*{References}
\noindent
Davis, Ernest (2013). 
The relevance of proofs of the rationality of probability theory
to automated reasoning and cognitive models. Unpublished. \\
\url{https://cs.nyu.edu/faculty/davise/papers/ProbProofs.pdf}

\noindent
Kehler, Andrew. (2005). Testing for Common Sense: Thoughts on Pronoun 
Interpretation and the Winograd Schema Challenge.
Workshop on Language \& Common Sense: Integrating Across Psychology,
Linguistics, and Computer Science {\em CogSci-2015}.

\noindent
Kocijan, Vid, Thomas Lukasiewicz, Ernest Davis, Gary Marcus, and 
Leora Morgenstern (2020). The defeat of the Winograd Schema Challenge.
arXiv preprint arXiv:2004.13831; submitted to {\em Artificial Intelligence}.
\url{https://arxiv.org/abs/2004.13831}

\noindent
Lake, Brenden M., and Gregory L. Murphy (2020). 
Word meaning in minds and machines. arXiv preprint arXiv:2008.01766.
\url{https://arxiv.org/abs/2008.01766}

\noindent
Levesque, Hector, Ernest Davis, and Leora Morgenstern (2012). The Winograd
schema challenge. {\em KR-2012.}

\noindent
Rohde, Hannah. 2018. Pronoun interpretation and production. 
In Chris Cummins and Napoleon Katsos (eds.) 
{\em The Oxford Handbook of Experimental Semantics and Pragmatics.}
Oxford University Press.

\noindent
Sakaguchi, Keisuke, Ronan Le Bras, Chandra Bhagavatula, and Yejin 
Choi (2020). Winogrande: An adversarial Winograd schema challenge at scale. 
In {\em Proceedings of the AAAI Conference on Artificial Intelligence,} 
pp.~8732-8740.

\end{document}